\documentclass[letterpaper, 10 pt, conference]{ieeeconf}  %

\IEEEoverridecommandlockouts                              %

\usepackage{multirow}
\usepackage{graphicx}
\usepackage{booktabs}
\usepackage{multirow}
\usepackage{graphicx}
\usepackage[table]{xcolor}
\usepackage{longtable}
\usepackage{booktabs}
\usepackage{adjustbox} %
\usepackage[margin=1in]{geometry}
\usepackage{amsmath,amssymb}
\usepackage{graphicx}
\usepackage{indentfirst}
\usepackage{url}
\usepackage{breakurl}
\usepackage{hyperref}

\title{\LARGE \bf
AutoMisty: A Multi-Agent LLM Framework for Automated Code Generation in the Misty Social Robot}

\author{Xiao Wang$^{1,\dag}$, Lu Dong$^{1, \dag}$, Sahana Rangasrinivasan$^{1,2}$, \\ Ifeoma Nwogu$^{1}$, Srirangaraj Setlur$^{1}$, Venugopal Govindaraju$^{1}$%
\thanks{\dag \quad Equal Contribution}%
\thanks{$^{1}$ Authors are with the State University of New York at Buffalo, Buffalo, NY 14260, USA. Emails: \{xwang277, ludong, srangasr, inwogu, setlur, govind\}@buffalo.edu
        }%
\thanks{$^{2}$Author is with the Department of Computer Science and Engineering, Amrita School of Computing, Amrita Vishwa Vidyapeetham, Amritapuri, Kollam, 690525, Kerala, India.
        }%
}

\begin{document}

\maketitle
\thispagestyle{empty}
\pagestyle{empty}

\begin{abstract}

The social robot's open API allows users to customize open-domain interactions. However, it remains inaccessible to those without programming experience. We introduce AutoMisty, the first LLM-powered multi-agent framework that converts natural-language commands into executable Misty robot code by decomposing high-level instructions, generating sub-task code, and integrating everything into a deployable program. Each agent employs a two-layer optimization mechanism: first, a self-reflective loop that instantly validates and automatically executes the generated code, regenerating whenever errors emerge; second, human review for refinement and final approval, ensuring alignment with user preferences and preventing error propagation. To evaluate AutoMisty's effectiveness, we designed a benchmark task set spanning four levels of complexity and conducted experiments in a real Misty robot environment. Extensive evaluations demonstrate that AutoMisty not only consistently generates high-quality code but also enables precise code control, significantly outperforming direct reasoning with ChatGPT-4o and ChatGPT-o1. All code, optimized APIs, and experimental videos will be publicly released through the webpage:  
\href{https://wangxiaoshawn.github.io/AutoMisty.html}{AutoMisty}.

\end{abstract}

\section{INTRODUCTION}

Social robots are becoming increasingly prevalent in homes, workplaces, and educational settings, providing companionship, assistance, and engagement across various application \cite{matheus2025long}. Misty, a versatile social robot, is gaining adoption for interactive and assistive roles. For example, it can serve as a mental well-being coach \cite{spitale2023robotic}, a caregiver for the elderly \cite{ciuffreda2023design}, or a learning companion for children \cite{cagiltay2022understanding}. As these robots evolve to support diverse interactions, customizing their behavior to align with individual preferences remains a challenge. Despite the availability of open APIs and developer-friendly frameworks, personalization often requires advanced programming skills, making it inaccessible to non-technical users \cite{mower2024ros, liang2022iropro}.

Large Language Models (LLMs) are emerging as a promising solution for enhancing human-robot interaction due to their advanced code generation capabilities. Recent research has explored leveraging LLMs to improve programming systems, reducing the need for extensive coding skills \cite{ge2024cocobo, swaminathan2024if}. However, a fully automated, end-to-end system has yet to be realized.
At the same time, existing LLM-driven frameworks face significant challenges, including reasoning limitations, hallucinations, and execution reliability issues, leading to frequent errors in robot programming and action execution \cite{chen2023forgetful}. Additionally, the absence of robust feedback mechanisms often leads to undesired task behavior, resulting in error propagation. This issue is further exacerbated by the context length limitations of many large language models, which struggle to retain essential information throughout prolonged interactions \cite{maharana2024evaluating}.
Recent work in robotics has explored multi-agent collaboration systems based on large language models (LLMs), incorporating structured reasoning, task planning, and dedicated agent collaboration to improve task execution reliability \cite{kannan2024smart}. However, further refinement is needed to ensure adaptability and responsiveness in fully end-to-end systems.

\begin{figure}[t]
    \centering
    \includegraphics[width=1\linewidth]{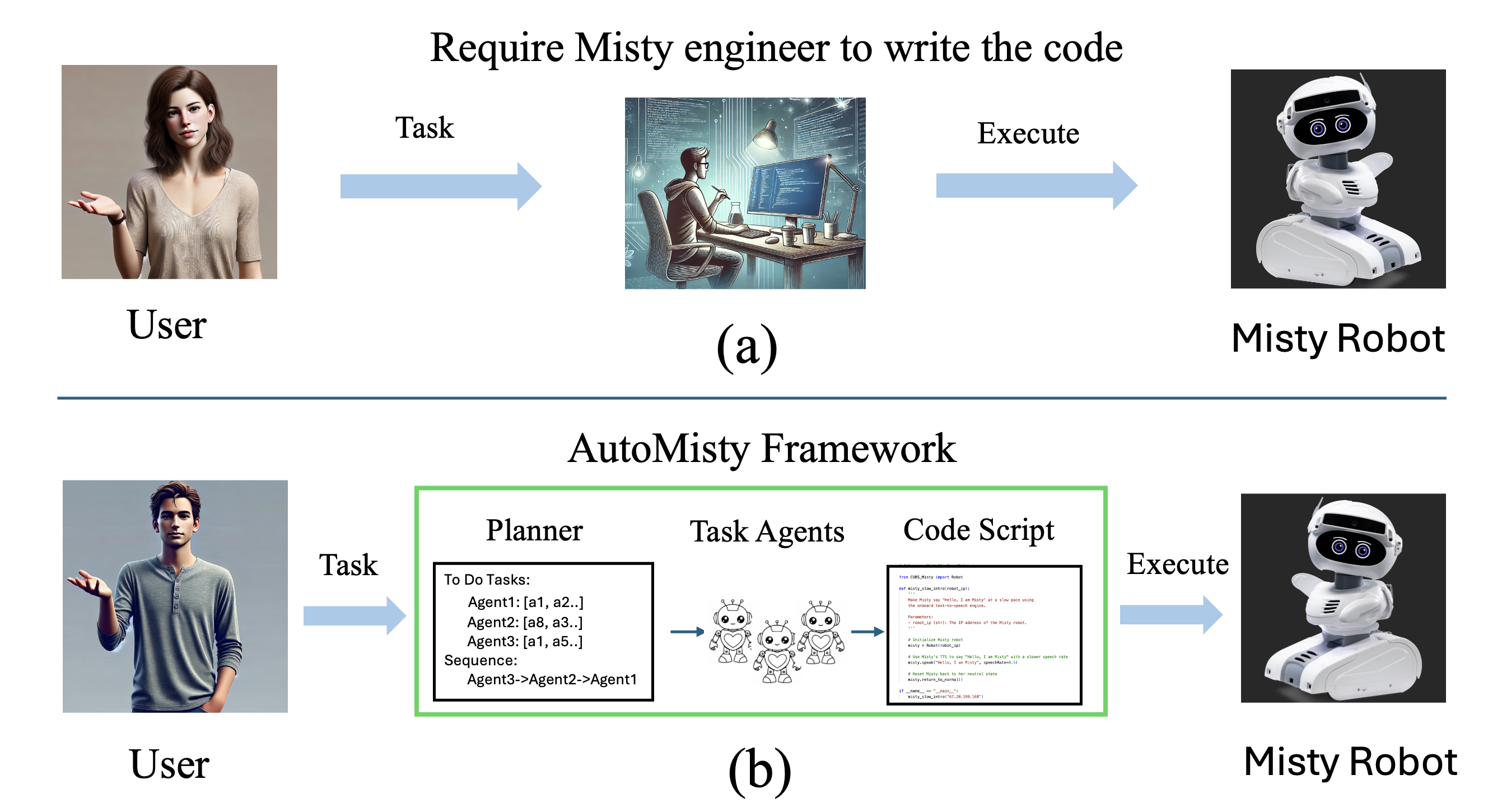}
    \caption{ An overview of the workflow comparison: (a) the previous way to customizing human-robot interaction and (b) the AutoMisty framework for enhanced customization. AutoMisty enables autonomous task planning, task assignment, self-reflection, and executable code generation while adapting to user preferences.}
    \label{fig:overview}
\end{figure}

To tackle these challenges, we introduce AutoMisty (illustrated in Fig.\ref{fig:overview}), a novel multi-agent LLM framework designed to generate reliable Misty robot control code from natural language task descriptions. The framework supports abstract task understanding and planning, subtask assignment, executable code generation, and real-time code deployment to the Misty robot. Furthermore, users can provide feedback based on the robot’s actual performance, enabling human-in-the-loop collaboration to improve code quality and execution outcomes. Our key contributions are as follows:

\begin{itemize}
\item  We present AutoMisty, the first multi-agent system designed for natural language-driven Misty robot programming. It autonomously formulates task plans, performs reasoning, and generates executable code, allowing Misty to execute complex instructions with user-specific adaptability.

\item We design a two-layer mechanism to ensure reliable and user-aligned task execution: (i) a self-reflective loop that validates and executes generated code in real time, automatically regenerating it when errors are detected; and (ii) a human feedback loop for iterative refinement and final approval, ensuring alignment with user intent and preventing error propagation. In addition, we optimize the Misty control APIs to improve interpretability and call reliability for LLMs.

\item  We construct a 28-task benchmark spanning different complexity levels and introduce five evaluation metrics. Extensive experiments and ablation studies showcase the effectiveness and efficiency of AutoMisty.
\end{itemize}

\begin{figure*}[t]
    \centering
    \includegraphics[width=0.9\linewidth]{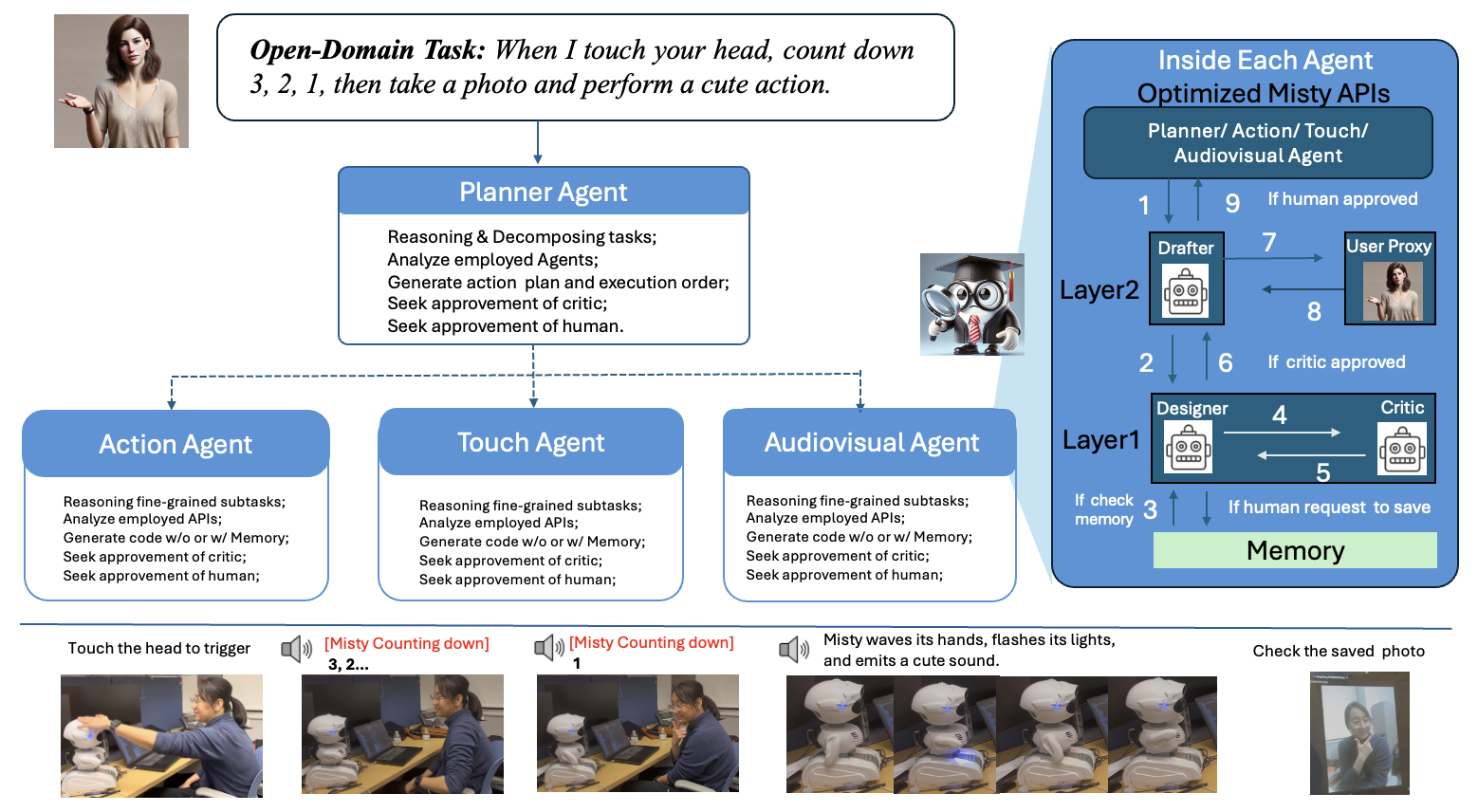}
    \caption{Overview of the AutoMisty Framework. Given an open-domain task description, the Planner agent decomposes the task, formulates a plan, and determines the execution order. It then routes subtasks to specialized agents (Action, Touch, or Audiovisual). Leveraging the optimized Misty API, each subtask agent refines details and generates executable code. Each agent incorporates a two-layer optimization mechanism: Layer 1 involves self-reflection with iterative feedback from a designer-critic interaction, while Layer 2 integrates human-in-the-loop feedback for continuous adaptation. Memory ensures consistency across future generations. The bottom illustrates the interaction process for this task.}
    \label{fig:automisty_overview}
\end{figure*}

\section{RELATED WORKS}

\noindent \textbf{AI for Code Generation: }Recent AI advancements enable automated code synthesis from natural language. State-of-the-art methods enhance performance through planning mechanisms \cite{zhang2023planning} or code-document retrieval \cite{zhou2023docprompting}. Multi-turn synthesis, as in CodeGen \cite{nijkamp2023codegen}, improves iteration but introduces challenges: reliance on predefined test cases limits generalization, documentation retrieval depends on external sources, and multi-turn refinement can compound errors. AutoMisty addresses these issues by integrating real-time user feedback and a multi-agent system, where specialized agents validate intermediate outputs, ensuring robustness, reducing cascading errors, and enhancing adaptability.

\noindent \textbf{LLM agents for Planning and Reasoning:}
Large language models (LLMs) serve as powerful tools for structured planning and reasoning. For instance, the self-consistency approach \cite{wang2022self} suggests repeatedly generating potential solutions and using majority voting to finalize the answer, while ReAct \cite{yao2023react} employs explicit reasoning and grounding to further enhance an LLM’s capacity for problem sovling. Iterative optimization techniques—such as self-explanation \cite{wang2023describe}, language reinforcement \cite{shinn2023reflexion}, tree of thought \cite{yao2023tree}, and chain of thought \cite{wei2022chain}—build upon past decisions to refine learning outcomes. Concurrently, state-based modeling \cite{wu2024stateflow}, curriculum-driven learning \cite{wang2024voyager}, and generative agents \cite{joon2023generative} promote adaptive reasoning, collectively driving progress in task planning. Drawing inspiration from these methods, Automisty integrates human-in-the-loop validation \cite{wu2024stateflow} to ensure the reliability of outputs and mitigate error propagation. Furthermore, it introduces a Critic module for automated reflection \cite{shinn2023reflexion} and continual optimization, streamlining collaborative workflows.

\noindent \textbf{LLMs for Robotic Control:} LLMs have demonstrated significant potential in robotic control by leveraging vision-language-action models and interactive programming. Key advancements include repurposing code-generation models for robot policy synthesis \cite{liang2022code, ahn2022can, singh2023progprompt, huang2023grounded}, extracting actionable knowledge without retraining \cite{pmlr-v162-huang22a, li2022pre, huang2022inner, vemprala2024chatgpt, song2023llm}, integrating foundation models for hierarchical problem-solving \cite{ajay2023compositional}, and incorporating embodied language models (ELMs) with continuous sensor data \cite{driess2023palm}. However, these approaches either suffer from hallucination errors or require extensive pretraining. Automisty addresses these challenges through interactive programming, where a compiler verifies code correctness, and human observation provides real-world feedback on Misty's performance, effectively eliminating hallucination issues. Additionally, the system employs in-context learning, circumventing the need for model retraining.

\section{PROBLEM FORMULATION}

Given a natural language task \(\mathcal{I}\), AutoMisty aims to generate executable MistyCode code $C$ to accomplish the given task. Initially, the system performs semantic analysis and requirement delineation on \(\mathcal{I}\), then breaks it down into several sub-tasks $T = T_1, T_2, ..., T_k$. If a sub-task is too complex, it can be further subdivided until each sub-task is sufficiently atomic to execute.  The system then leverages a set of \( N \) LLM agents, denoted as \( A = \{A_1, A_2, \ldots, A_N\} \),  where each \( A_i \) specializes in handling a specific type of sub-task. An assignment function then maps each sub-task to its corresponding agent, whose capabilities match the subtask’s requirements. This is formally defined as:
\[
  \phi : \{\, T_1, \dots, T_k \} \;\to\; \{A_1, A_2, \dots, A_N\}
\]

The overall task can be naturally formulated as a directed acyclic graph (DAG), where agents are nodes and dependency relations define edges. If the output of agent \(A_j\) is required as input for agent \(A_i\), a directed edge \((A_j \to A_i)\) is added to a directed acyclic graph \(G\), which represents this multi-agent collaboration. As \(G\) is acyclic, a topological sort produces the execution sequence:
\[
  \pi(G) \;=\; \langle A_{i_1},\, A_{i_2},\, \dots,\, A_{i_m}\rangle,
\]
ensuring that for any agent in the sequence, all of its upstream dependencies have finished before it executes. For the \(k\)-th agent in the execution order, \(A_{i_k}\), its input includes the upstream-generated code \(\Delta(A_{i_k})\) and its own LLM context \(\text{Ctx}(A_{i_k})\), such as system prompts and subtask descriptions. The code generation function is defined as:
\[
  \Gamma(A_{i_k} \,\mid\, \text{Ctx}(A_{i_k}),\, \Delta(A_{i_k})) \;=\; C_{i_k},
\]
where \(C_{i_k}\) is the code generated by \(A_{i_k}\) for its assigned subtask. The final Misty Code output \(C\) corresponds to the output of the last agent in the topological execution sequence, completing the original task \(\mathcal{I}\).

\renewcommand{\arraystretch}{1.5}
\begin{table*}[]
    \caption{List of tasks under each category}
    \label{tab:task_list}
    \centering
    \begin{tabular}{|p{3cm}|p{10cm}|}
        \hline
    
        \textbf{Elementary Task} & (1) Hand Raise Direction Change, (2) Hand Raise Count Change, (3) Hand Raise Speed Change, (4) Head Turn Direction Change, (5) Head Turn Count Change, (6) Head Turn Speed Change, (7) Expression Transformation Test, (8) Speech Speed Change Test, (9) LED Color Change Test \\
        \hline
        
        \textbf{Simple Task} & (10) Storytelling, (11) Timed Reminder, (12) Emotional Comfort, (13) Poetry Composition and Performance, (14) Key Word Triggered Video Recording, (15) Rock-style Dance Performance, (16) Key Word Triggered Photography\\
        
        \hline
        \textbf{Compound Task} & (17) Reacting to Head Touch with Pikachu Noise, (18) Expressing Different Emotions Based on Touch Location, (19) Recognizing Colors and Displaying with LED, (20) Waving in Response to a Greeting\\
        
        \hline
        \textbf{Complex Task} & (21) Recognizing and Reading Signs, (22) Simulating a Child’s Persona in Conversation, (23) Math Tutoring, (24) Intelligent Storytelling, (25) Real-Time Translation, (26) Taking a Photo and Performing Analysis, (27) Countdown Before Taking a Photo and Performing an Action (28) Playing a Game Like Concentration\\
        \hline
    \end{tabular}
\end{table*}

\section{METHODOLOGY}

This section outlines the AutoMisty methodology. We first present the optimized Misty API design and its functional partitioning across agents. This design enhances LLM comprehension, mitigates context window issues that prevent accurate API selection and internal code modification, and helps agents focus on their assigned subtasks. We then introduce a unified two-layer architecture shared by all agents to ensure reliable and well-structured code generation. Finally, we describe the collaboration among the four core agents and their respective roles.Fig. \ref{fig:automisty_overview} illustrates this approach.

\subsection{Optimized Misty APIs}

We identified several issues in Misty’s original APIs, such as vague descriptions, missing parameters, and incomplete interfaces, often causing execution errors or hallucinations. To enable high-quality code generation, we optimized the API set. Following insights from Song et al. \cite{song2024code}, we found that detailed annotations significantly improve LLM performance. We analyzed Misty’s core capabilities—action, touch, and audiovisual processing—and restructured the APIs accordingly. This includes refining audio-video flow, integrating Whisper for real-time translation, and enabling basic verbal reasoning via LLM/VLM. The final set includes 136 optimized APIs with clear documentation on I/O formats, parameters, use cases, and complex functions. These APIs are embedded in system prompts and routed to agents for task-specific code generation.

\subsection{Two-layer Optimization Mechanism}

AutoMisty integrates a Two-Layer Optimization Mechanism within each agent to enhance reasoning, decision-making, and task execution. Upon activation, the agent triggers a Drafter that manages Layer 1, which adopts a Self-Reflective Feedback Mechanism: the Designer generates initial code via In-Context Learning (ICL) and iteratively refines it based on feedback from the Critic and, when available, the user. The Critic evaluates the code’s correctness and feasibility. If the solution is unsatisfactory, the Designer and Critic engage in a feedback loop until the Critic approves the optimized version. These interactions are orchestrated through sequential prompting, enabling structured, stepwise refinement within the LLM context.

The Designer also incorporates a memory module based on retrieval-augmented generation (RAG). When users save a successful task, the system stores the verified code and preferences. For future similar tasks, the module retrieves relevant examples before generation, providing few-shot demonstrations that align with user intent and prior successes.

After Layer~1, the Drafter proceeds to Layer~2---the human-in-the-loop phase. The UserProxy component integrates a compiler tailored to the Misty runtime environment and automatically executes the generated code. If execution errors occur, they are appended to the context and returned to the Designer for revision. Once the code runs successfully and is deployed on the Misty robot, the user can provide feedback on its behavior via typed input. If the user is not satisfied, the Drafter reactivates Layer 1 and incorporates the human feedback into the context to further refine the output. The subtask is only considered complete after the user explicitly confirms success.

\subsection{Multi-agent Collaboration Framework}

AutoMisty's multi-agent collaboration framework consists of four key agents: the Planner agent, Action agent, Touch agent, and Audiovisual agent, each contributing to different aspects of task execution while ensuring iterative validation and refinement.

The \textbf{Planner agent} serves as the central coordinator, interpreting vague high-level human instructions, enriching them with necessary details, decomposing tasks into manageable subtasks, and assigning them to the appropriate agents.It generates a coherent, optimized execution plan, validates it with the user, and routes subtasks to agents accordingly.

\textbf{Action, Touch, and Audiovisual agents} each operate within their respective functional domains. The Action agent controls Misty’s speech output, gestures, and head movements. The Touch agent processes physical interactions from users and automatically translates them into system-recognizable responses. The Audiovisual agent handles visual and auditory inputs, integrating both a large language model (LLM) and a vision-language model (VLM), enabling Misty to reason and respond based on what it sees and hears, thereby providing fundamental multimodal understanding and interaction capabilities.

In addition, each agent is designed to access its own internal context and is unaware of the intermediate reasoning processes of other agents. At most, it receives the final, verified code output from the preceding agent, which must be embedded \textbf{verbatim} into its own generated code. This design prevents context overflow caused by long cross-agent context histories and ensures that each agent remains focused solely on its assigned subtask, thereby improving both efficiency and system robustness.

\begin{table*}[t].
\centering          
\caption{Performance Comparison of ChatGPT-4o, ChatGPT-o1, and AutoMisty Across Different Categories of Tasks }
\label{tab:comparison_with_baselines}
\renewcommand{\arraystretch}{1.2} %
\centering
\begin{tabular}{%
    |c  %
    ||c|c|c|c|c  %
    ||c|c|c|c|c  %
    ||c|c|c|c|c| %
}
\hline
\multirow{3}{*}{\textbf{Tasks}} 
& \multicolumn{5}{c||}{\textbf{ChatGPT\_4o}} 
& \multicolumn{5}{c||}{\textbf{ChatGPT\_o1}} 
& \multicolumn{5}{c|}{\textbf{AutoMisty (Ours)}} 
\\
\cline{2-16}
& \textbf{TC}
& \multicolumn{2}{c|}{\textbf{NUI}}
& \textbf{CE} & \textbf{US}
& \textbf{TC}
& \multicolumn{2}{c|}{\textbf{NUI}}
& \textbf{CE} & \textbf{US}
& \textbf{TC}
& \multicolumn{2}{c|}{\textbf{NUI}}
& \textbf{CE} & \textbf{US}
\\
\cline{3-4} \cline{8-9} \cline{13-14}
& 
& \textbf{UPI} & \textbf{TCI}
& 
& 
& 
& \textbf{UPI} & \textbf{TCI}
& 
& 
& 
& \textbf{UPI} & \textbf{TCI}
& 
& 
\\
\hline
\multicolumn{16}{|c|}{\textbf{Elementary Tasks}} \\
\hline
1 - 4  & Success & 0 & 0 & 10 & 10 & Success & 0 & 0 & 10 & 10 & Success & 0 & 0 & 10 & 10 \\
\hline
5  & Success & 1 & 0 & 10 & 10 & Success & 0 & 1 & 10 & 10 & Success & 1 & 0 & 10 & 10 \\
\hline
6  & Fail    & - & - & -  & -  & Success & 0 & 0 & 10 & 10 & Success & 0 & 0 & 10 & 10 \\
\hline
7  & Success & 2 & 1 & 10 & 10 & Success & 0 & 1 & 10 & 10 & Success & 1 & 0 & 10 & 10 \\
\hline
8  & Success & 0 & 0 & 10 & 10 & Success & 0 & 0 & 10 & 10 & Success & 0 & 0 & 10 & 10 \\
\hline
9  & Success & 0 & 0 & 10 & 10 & Success & 0 & 1 & 10 & 10 & Success & 1 & 0 & 10 & 10 \\
\hline
\multicolumn{16}{|c|}{\textbf{Simple Tasks}} \\
\hline
10 & Success & 1 & 0 & 10 & 10 & Success & 1 & 0 & 10 & 10 & Success & 1 & 0 & 10 & 10 \\
\hline
11 & Success & 0 & 0 & 10 & 10 & Success & 0 & 0 & 10 & 10 & Success & 0 & 0 & 10 & 10 \\
\hline
12 & Success & 1 & 0 & 10 & 10 & Success & 0 & 0 & 10 & 10 & Success & 1 & 0 & 10 & 10 \\
\hline
13 & Success & 0 & 0 & 10 & 10 & Success & 0 & 0 & 10 & 10 & Success & 0 & 0 & 10 & 10 \\
\hline
14 & Fail    & - & - & -  & -  & Fail    & - & - & - & - & Success & 0 & 0 & 10 & 10 \\
\hline
15 & Success & 0 & 0 & 10 & 10 & Success & 0 & 0 & 10 & 10 & Success & 0 & 0 & 10 & 10 \\
\hline
16 & Fail    & - & - & -  & -  & Success & 0 & 0 & 10 & 10 & Success & 0 & 0 & 10 & 10 \\
\hline
\multicolumn{16}{|c|}{\textbf{Compound Tasks}} \\
\hline
17 & Fail    & - & - & -  & -  & Success & 1 & 0 & 10 & 10 & Success & 0 & 1 & 10 & 10 \\
\hline
18 & Fail    & - & - & -  & -  & Fail    & - & - & -  & -  & Success & 1 & 0 & 10 & 10 \\
\hline
19 & Fail    & - & - & -  & -  & Success & 1 & 0 & 10 & 10 & Success & 1 & 0 & 9  & 10 \\
\hline
20 & Fail    & - & - & -  & -  & Success & 1 & 0 & 10 & 10 & Success & 1 & 0 & 9  & 10 \\
\hline
\multicolumn{16}{|c|}{\textbf{Complex Tasks}} \\
\hline
21 & Fail    & - & - & -  & -  & Success & 0 & 0 & 10 & 10 & Success & 0 & 1 & 9 & 10 \\
\hline
22 & Success & 0 & 0 & 10 & 10 & Success & 0 & 0 & 10 & 10 & Success & 0 & 0 & 9 & 10 \\
\hline
23 & Fail    & - & - & -  & -  & Success & 0 & 0 & 10 & 10 & Success & 0 & 0 & 9 & 10 \\
\hline
24 & Success    & 0 & 0 & 9  & 10  & Success & 0 & 0 & 10 & 10 & Success & 0 & 0 & 10 & 10 \\
\hline
25 & Fail    & - & - & -  & -  & Success & 0 & 0 & 10 & 10 & Success & 0 & 0 & 9 & 10 \\
\hline
26 & Fail    & - & - & -  & -  & Fail & - & - & - & - & Success & 0 & 0 & 9 & 10 \\
\hline
27 & Fail    & - & - & -  & -  & Fail & - & - & - & - & Success & 1 & 2 & 10 & 10 \\
\hline
28 & Fail    & - & - & -  & -  & Success & 6 & 4 & 10 & 8  & Success & 3 & 2 & 9 & 10 \\
\hline
\end{tabular}
\end{table*}

\section{EXPERIMENTS}

\subsection{Benchmark Dataset}

To evaluate AutoMisty’s performance, we constructed a natural language task suite for the Misty robot. Tasks are freely described without constraints on format or detail, allowing highly abstract instructions (e.g., “do a rock dance”). We collected task proposals from 10 participants (3 engineers, 7 non-engineers), merged overlapping entries, and asked each participant to define their own success criteria. The final suite covers tasks of varying complexity, grouped into four categories.

\begin{itemize}

    \item \textbf{Elementary Tasks} are basic low-level actions executable via a single API call, such as raising a hand, adjusting head orientation, or setting speed. They evaluate the system’s ability to map natural language commands to atomic robot controls.

    \item \textbf{Simple Tasks} are tasks that can be completed by a single agent, with clear solutions and straightforward execution.

    \item \textbf{Compound Tasks} require collaboration among multiple agents. The instructions are clear and Misty can complete the task independently without relying on an LLM as the central reasoning component.

    \item \textbf{Complex Tasks} involve abstract and ambiguous instructions that require coordination among multiple agents and rich human-robot interaction. The system must autonomously design the prompts for the LLM or VLM acting as Misty’s cognitive core. Solving these tasks demands strong abstraction understanding and planning capabilities.
    
\end{itemize}

The task suite consists of 28 tasks in total: 9 elementary tasks, 7 simple tasks, 4 compound tasks, and 8 complex tasks. Table \ref{tab:task_list} provides a list of tasks under each category.

\begin{table*}[!t]
\centering
\caption{Performance Comparison of AutoMisty With vs.\ Without Self-Reflection}
\label{tab:reflection_experiments}
\renewcommand{\arraystretch}{1.2}
\begin{tabular}{%
    |c                         %
    ||c|c|c|c|c                %
    ||c|c|c|c|c|               %
}
\hline
\multirow{3}{*}{\textbf{Tasks}} 
& \multicolumn{5}{c||}{\textbf{AutoMisty (W/O Self-Reflection)}}
& \multicolumn{5}{c|}{\textbf{AutoMisty (W Self-Reflection)}}
\\
\cline{2-11}
& \textbf{TC}
& \multicolumn{2}{c|}{\textbf{NUI}}
& \textbf{CE}
& \textbf{US}
& \textbf{TC}
& \multicolumn{2}{c|}{\textbf{NUI}}
& \textbf{CE}
& \textbf{US}
\\
\cline{3-4} \cline{8-9}
& 
& \textbf{UPI}
& \textbf{TCI}
& 
& 
& 
& \textbf{UPI}
& \textbf{TCI}
& 
& 
\\
\hline
\multicolumn{11}{|c|}{\textbf{Elementary Tasks}} \\
\hline
1 - 9 & Success & 0 & 0 & 10 & 10 & Success & 0 & 0 & 10 & 10 \\
\hline
\multicolumn{11}{|c|}{\textbf{Simple Tasks}} \\
\hline
10 & Success & 1 & 0 & 10 & 10 & Success & 1 & 0 & 10 & 10 \\
\hline
11 & Success & 1 & 0 & 10 & 10 & Success & 1 & 0 & 10 & 10 \\
\hline
12 & Success & 0 & 0 & 10 & 10 & Success & 0 & 0 & 10 & 10 \\
\hline
13 & Success & 0 & 2 & 10 & 10 & Success & 0 & 0 & 10 & 10 \\
\hline
14 & Success & 0 & 0 & 10 & 10 & Success & 0 & 0 & 10 & 10 \\
\hline
15 & Success & 0 & 0 & 10 & 10 & Success & 0 & 0 & 10 & 10 \\
\hline
16 & Success & 0 & 0 & 10 & 10 & Success & 0 & 0 & 10 & 10 \\
\hline
\multicolumn{11}{|c|}{\textbf{Compound Tasks}} \\
\hline
17 & Success & 0 & 0 & 10 & 6  & Success & 0 & 1 & 10 & 5  \\
\hline
18 & Success & 0 & 0 & 10 & 10 & Success & 0 & 1 & 10 & 10 \\
\hline
19 & Success & 1 & 1 & 10 & 10 & Success & 1 & 0 & 9  & 10 \\
\hline
20 & Success & 0 & 0 & 10 & 10 & Success & 1 & 0 & 9  & 10 \\
\hline
\multicolumn{11}{|c|}{\textbf{Complex Tasks}} \\
\hline
21 & Fail    & - & - & - & - & Success & 0 & 1 & 9  & 10 \\
\hline
22 & Fail    & - & - & - & - & Success & 0 & 0 & 9  & 10 \\
\hline
23 & Success & 0 & 0 & 10 & 10   & Success & 0 & 0 & 9  & 10 \\
\hline
24 & Success & 0 & 0 & 10 & 10   & Success & 0 & 0 & 9  & 10 \\
\hline
25 & Success & 0 & 0 & 10 & 10   & Success & 0 & 0 & 9  & 10 \\
\hline
26 & Success & 0 & 0 & 10 & 10   & Success & 0 & 0 & 9  & 10 \\
\hline
27 & Success & 0 & 0 & 10 & 10   & Success & 1 & 2 & 10 & 10 \\
\hline
28 & Fail    & - & - & - & - & Success & 3 & 2 & 9  & 10 \\
\hline
\end{tabular}
\end{table*}

\subsection{Evaluation Metrics}

We evaluate performance using the following metrics: Task Completion (TC) status, Number of User Interactions (NUI), Code Efficiency (CE) score, and User Satisfaction (US) rating. Within NUI, we further assess User Preference Interactions (UPI) and Technical Correctness Interactions (TCI).

\begin{itemize}

    \item \textit{TC} indicates whether the task is completed within 20 interactions.

    \item \textit{NUI} represents the total number of user interactions during the process.

    \item \textit{UPI} counts interactions where the user participated in an interaction due to personal preference.

    \item \textit{TCI} tracks interactions where the user provides suggestions related to coding correctness or implementation logic.

    \item \textit{CE} evaluates the efficiency of the generated code. If a portion of the code can be removed without affecting correctness, a penalty of -1 is applied. The maximum possible score is 10.

    \item \textit{US} measures user satisfaction with the generated code. Users were asked to document their interactions while using AutoMisty. If the system misinterpreted or failed to follow an instruction, a penalty of 1 point was applied. The maximum score for this metric is 10.
    
\end{itemize}

\section{RESULTS AND DISCUSSION}

This section presents experimental results with AutoMisty and baseline methods on unseen dataset tasks, as summarized in Table \ref{tab:comparison_with_baselines}.

We observe that for elementary and simple tasks, all models achieve a high success rate with minimal user interaction. However, as task complexity increases, we find that AutoMisty and ChatGPT-o1 begin to outperform ChatGPT-4o. The most significant differences emerge in complex tasks, where AutoMisty shows the highest robustness and adaptability. ChatGPT-4o failed most complex tasks (4 out of 8) while ChatGPT-o1 failed a few (2 out of 8). Although AutoMisty's CE is a little lower than that of ChatGPT-o1, it was able to complete all tasks successfully. This demonstrates AutoMisty's capability to generate MistyCode across all task complexities.

\noindent \textbf{Performance Variation Analysis:} The non-deterministic behavior of LLMs leads to inherent variability in their outputs \cite{Ouyang_2025}. To analyze this, we performed five independent trials, selecting the tasks that both baseline models, ChatGPT-4o and ChatGPT-o1, failed to complete. Table \ref{tab:per-task-no-num-part} reports the mean and standard deviation of the results obtained across these trials for our approach. We observe that our proposed method consistently produces similar outcomes across all tasks.

\begin{table}[ht]
\centering
\caption{AutoMisty Performance Variation Analysis}
\label{tab:per-task-no-num-part}
\resizebox{\columnwidth}{!}{
\begin{tabular}{lccccc}
\hline
\textbf{Task} & 
\textbf{TC} & 
\textbf{UPI} & 
\textbf{TCI} & 
\textbf{CE} & 
\textbf{US} \\
\hline
14 &
1.00$\pm$0.00 &
0.00$\pm$0.00 &
0.00$\pm$0.00 &
10.00$\pm$0.00 &
10.00$\pm$0.00 \\

18 &
1.00$\pm$0.00 &
0.00$\pm$0.00 &
0.00$\pm$0.00 &
10.00$\pm$0.00 &
10.00$\pm$0.00 \\

26 &
1.00$\pm$0.00 &
0.00$\pm$0.00 &
0.00$\pm$0.00 &
10.00$\pm$0.00 &
10.00$\pm$0.00 \\

27 &
1.00$\pm$0.00 &
0.00$\pm$0.00 &
1.00$\pm$1.00 &
10.00$\pm$0.00 &
10.00$\pm$0.00 \\
\hline
\end{tabular}
}
\end{table}

\begin{table*}[ht]
\centering
\caption{Teachability  Assessment}
\label{tab:emotion_rag}
\renewcommand{\arraystretch}{1.2} %
\begin{tabular}{|c|c|c|}
\hline
\textbf{Emotion Saved in Memory} & \textbf{User Input} & \textbf{Retrieval Result (Teachability Obseved)} \\
\hline
\multirow{3}{*}{Happiness and Excitement} 
    & Exhilaration & Happiness and Excitement (correct) \\
    & Ecstasy & Happiness and Excitement (correct) \\
    & Jubilation & Peaceful Happiness and Satisfaction (incorrect) \\
\hline
\multirow{3}{*}{Rage} 
    & Fury & Rage (correct) \\
    & Wrath & Rage (correct) \\
    & Outrage & Rage (correct) \\
\hline
\multirow{3}{*}{Grief} 
    & Sadness & Grief (correct) \\
    & Mourning & Grief (correct) \\
    & Heartache & Grief (correct) \\
\hline
\multirow{3}{*}{Peaceful Happiness and Satisfaction} 
    & Tranquil Bliss & Peaceful Happiness and Satisfaction (correct) \\
    & Quiet Euphoria & Happiness and Excitement (incorrect) \\
    & Placid Joy & Peaceful Happiness and Satisfaction (correct) \\
\hline
\end{tabular}
\end{table*}

\noindent \textbf{Impact of the Self-Reflective Feedback Mechanism:} To analyze the impact of the Self-Reflective Feedback Mechanism on AutoMisty's performance, we conduct experiments across all four task complexities without self-reflection. These results are highlighted in Table \ref{tab:reflection_experiments}. We find that both versions of AutoMisty performed consistently well on both elementary and simple tasks, achieving perfect task completion rates with minimal or no user interactions. In contrast, AutoMisty without the mechanism failed on multiple occasions (Tasks 21, 22, and 28), while the Self-Reflective version was able to complete all tasks. Although Self-Reflection led to a slight increase in the number of interactions (NUI) in complex tasks, it consistently resulted in higher task success rates, demonstrating its overall effectiveness.

\noindent \textbf{Teachability  Assessment:} Teachability refers to an agent’s ability to learn from prior successes and adapt to user preferences. AutoMisty supports this by using a vector database for retrieval, where each key is a compressed subtask description ((5--10 word high-level summary generated via an LLM)), and the value is user-verified code. These examples reflect both prior successes and user preferences, guiding future generations toward more personalized outputs.

We evaluated this capability through an emotion-grounded experiment. We first injected 10 irrelevant tasks into the memory and stored four anchor emotions (e.g., Rage, Grief), each mapped to expressive Misty behaviors involving motion, facial expressions, and sound. Users then input synonymous or vague emotion terms. For each query, we retrieved only the top-1 result from memory. As shown in Table~\ref{tab:emotion_rag}, AutoMisty correctly retrieved the corresponding emotion class code for clearly defined inputs, demonstrating teachability. Additionally, the generated code for each emotion exhibited consistent behavioral patterns.

\section{CONCLUSION}

AutoMisty is the first system to break the technical bottleneck, enabling non-programmers to generate executable code directly through conversational instructions. Our design significantly improves code controllability, accuracy, and consistency, demonstrating strong robustness across tasks of varying complexity. Although the current task suite includes only 28 examples, the system exhibits strong generalization capabilities for broader open-domain tasks. The entire AutoMisty framework is built on in-context learning, enabling low-cost and efficient transfer to other API-driven social robotics platforms.

\section*{Acknowledgment}
This material is based upon work supported under the AI Research Institutes program by the U.S. National Science Foundation and the Institute of Education Sciences, U.S. Department of Education, through Award \# 2229873—National AI Institute for Exceptional Education. Any opinions, findings and conclusions, or recommendations expressed in this material are those of the author(s) and do not necessarily reflect the views of the National Science Foundation, the Institute of Education Sciences, or the U.S. Department of Education.

\addtolength{\textheight}{0cm}   %

\scriptsize
\bibliographystyle{IEEEtran} %
\bibliography{IEEEexample} %

\begin{thebibliography}{10}
\providecommand{\url}[1]{#1}
\csname url@rmstyle\endcsname
\providecommand{\newblock}{\relax}
\providecommand{\bibinfo}[2]{#2}
\providecommand\BIBentrySTDinterwordspacing{\spaceskip=0pt\relax}
\providecommand\BIBentryALTinterwordstretchfactor{4}
\providecommand\BIBentryALTinterwordspacing{\spaceskip=\fontdimen2\font plus
\BIBentryALTinterwordstretchfactor\fontdimen3\font minus \fontdimen4\font\relax}
\providecommand\BIBforeignlanguage[2]{{%
\expandafter\ifx\csname l@#1\endcsname\relax
\typeout{** WARNING: IEEEtran.bst: No hyphenation pattern has been}%
\typeout{** loaded for the language `#1'. Using the pattern for}%
\typeout{** the default language instead.}%
\else
\language=\csname l@#1\endcsname
\fi
#2}}

\bibitem{matheus2025long}
K.~Matheus, R.~Ramnauth, B.~Scassellati, and N.~Salomons, ``Long-term interactions with social robots: Trends, insights, and recommendations,'' \emph{ACM Transactions on Human-Robot Interaction}, vol.~14, no.~3, pp. 1--42, 2025.

\bibitem{spitale2023robotic}
M.~Spitale, M.~Axelsson, and H.~Gunes, ``Robotic mental well-being coaches for the workplace: An in-the-wild study on form,'' in \emph{Proceedings of the 2023 ACM/IEEE International Conference on Human-Robot Interaction}, 2023, pp. 301--310.

\bibitem{ciuffreda2023design}
I.~Ciuffreda, G.~Amabili, S.~Casaccia, M.~Benadduci, A.~Margaritini, E.~Maranesi, F.~Marconi, A.~De~Masi, J.~Alberts, J.~de~Koning, \emph{et~al.}, ``Design and development of a technological platform based on a sensorized social robot for supporting older adults and caregivers: Guardian ecosystem,'' \emph{International Journal of Social Robotics}, pp. 1--20, 2023.

\bibitem{cagiltay2022understanding}
B.~Cagiltay, N.~T. White, R.~Ibtasar, B.~Mutlu, and J.~Michaelis, ``Understanding factors that shape children’s long term engagement with an in-home learning companion robot,'' in \emph{Proceedings of the 21st annual ACM interaction design and children conference}, 2022, pp. 362--373.

\bibitem{mower2024ros}
C.~E. Mower, Y.~Wan, H.~Yu, A.~Grosnit, J.~Gonzalez-Billandon, M.~Zimmer, J.~Wang, X.~Zhang, Y.~Zhao, A.~Zhai, \emph{et~al.}, ``Ros-llm: A ros framework for embodied ai with task feedback and structured reasoning,'' \emph{arXiv preprint arXiv:2406.19741}, 2024.

\bibitem{liang2022iropro}
Y.~S. Liang, D.~Pellier, H.~Fiorino, and S.~Pesty, ``iropro: An interactive robot programming framework,'' \emph{International Journal of Social Robotics}, vol.~14, no.~1, pp. 177--191, 2022.

\bibitem{ge2024cocobo}
Y.~Ge, Y.~Dai, R.~Shan, K.~Li, Y.~Hu, and X.~Sun, ``Cocobo: Exploring large language models as the engine for end-user robot programming,'' in \emph{2024 IEEE Symposium on Visual Languages and Human-Centric Computing (VL/HCC)}.\hskip 1em plus 0.5em minus 0.4em\relax IEEE, 2024, pp. 89--95.

\bibitem{swaminathan2024if}
M.~Swaminathan, L.-J. Hsu, M.~M. Thant, K.~J. Amon, A.~S. Kim, K.~M. Tsui, S.~Sabanovi{\'c}, D.~J. Crandall, and W.~Khoo, ``If [yourname] can code, so can you! end-user robot programming for non-experts,'' in \emph{Companion of the 2024 ACM/IEEE International Conference on Human-Robot Interaction}, 2024, pp. 1033--1037.

\bibitem{chen2023forgetful}
J.-T. Chen and C.-M. Huang, ``Forgetful large language models: Lessons learned from using llms in robot programming,'' in \emph{Proceedings of the AAAI Symposium Series}, vol.~2, no.~1, 2023, pp. 508--513.

\bibitem{maharana2024evaluating}
A.~Maharana, D.-H. Lee, S.~Tulyakov, M.~Bansal, F.~Barbieri, and Y.~Fang, ``Evaluating very long-term conversational memory of llm agents,'' \emph{arXiv preprint arXiv:2402.17753}, 2024.

\bibitem{kannan2024smart}
S.~S. Kannan, V.~L. Venkatesh, and B.-C. Min, ``Smart-llm: Smart multi-agent robot task planning using large language models,'' in \emph{2024 IEEE/RSJ International Conference on Intelligent Robots and Systems (IROS)}.\hskip 1em plus 0.5em minus 0.4em\relax IEEE, 2024, pp. 12\,140--12\,147.

\bibitem{zhang2023planning}
S.~Zhang, Z.~Chen, Y.~Shen, M.~Ding, J.~B. Tenenbaum, and C.~Gan, ``Planning with large language models for code generation,'' in \emph{The Eleventh International Conference on Learning Representations}, 2023.

\bibitem{zhou2023docprompting}
S.~Zhou, U.~Alon, F.~F. Xu, Z.~Jiang, and G.~Neubig, ``Docprompting: Generating code by retrieving the docs,'' in \emph{The Eleventh International Conference on Learning Representations}, 2023.

\bibitem{nijkamp2023codegen}
E.~Nijkamp, B.~Pang, H.~Hayashi, L.~Tu, H.~Wang, Y.~Zhou, S.~Savarese, and C.~Xiong, ``Codegen: An open large language model for code with multi-turn program synthesis,'' in \emph{The Eleventh International Conference on Learning Representations}, 2023.

\bibitem{wang2022self}
X.~Wang, J.~Wei, D.~Schuurmans, Q.~Le, E.~Chi, S.~Narang, A.~Chowdhery, and D.~Zhou, ``Self-consistency improves chain of thought reasoning in language models,'' \emph{arXiv preprint arXiv:2203.11171}, 2022.

\bibitem{yao2023react}
S.~Yao, J.~Zhao, D.~Yu, N.~Du, I.~Shafran, K.~Narasimhan, and Y.~Cao, ``React: Synergizing reasoning and acting in language models,'' in \emph{International Conference on Learning Representations (ICLR)}, 2023.

\bibitem{wang2023describe}
Z.~Wang, S.~Cai, G.~Chen, A.~Liu, X.~Ma, and Y.~Liang, ``Describe, explain, plan and select: Interactive planning with llms enables open-world multi-task agents,'' in \emph{Thirty-seventh Conference on Neural Information Processing Systems}, 2023.

\bibitem{shinn2023reflexion}
N.~Shinn, F.~Cassano, A.~Gopinath, K.~R. Narasimhan, and S.~Yao, ``Reflexion: language agents with verbal reinforcement learning,'' in \emph{Thirty-seventh Conference on Neural Information Processing Systems}, 2023.

\bibitem{yao2023tree}
S.~Yao, D.~Yu, J.~Zhao, I.~Shafran, T.~L. Griffiths, Y.~Cao, and K.~R. Narasimhan, ``Tree of thoughts: Deliberate problem solving with large language models,'' in \emph{Thirty-seventh Conference on Neural Information Processing Systems}, 2023.

\bibitem{wei2022chain}
J.~Wei, X.~Wang, D.~Schuurmans, M.~Bosma, F.~Xia, E.~Chi, Q.~V. Le, D.~Zhou, \emph{et~al.}, ``Chain-of-thought prompting elicits reasoning in large language models,'' \emph{Advances in neural information processing systems}, vol.~35, pp. 24\,824--24\,837, 2022.

\bibitem{wu2024stateflow}
Y.~Wu, T.~Yue, S.~Zhang, C.~Wang, and Q.~Wu, ``Stateflow: Enhancing {LLM} task-solving through state-driven workflows,'' in \emph{First Conference on Language Modeling}, 2024.

\bibitem{wang2024voyager}
G.~Wang, Y.~Xie, Y.~Jiang, A.~Mandlekar, C.~Xiao, Y.~Zhu, L.~Fan, and A.~Anandkumar, ``Voyager: An open-ended embodied agent with large language models,'' \emph{Transactions on Machine Learning Research}, 2024.

\bibitem{joon2023generative}
\BIBentryALTinterwordspacing
J.~S. Park, J.~C. O'Brien, C.~J. Cai, M.~R. Morris, P.~Liang, and M.~S. Bernstein, ``Generative agents: Interactive simulacra of human behavior,'' \emph{CoRR}, vol. abs/2304.03442, 2023. [Online]. Available: \url{https://doi.org/10.48550/arXiv.2304.03442}
\BIBentrySTDinterwordspacing

\bibitem{liang2022code}
J.~Liang, W.~Huang, F.~Xia, P.~Xu, K.~Hausman, brian ichter, P.~Florence, and A.~Zeng, ``Code as policies: Language model programs for embodied control,'' in \emph{Workshop on Language and Robotics at CoRL 2022}, 2022.

\bibitem{ahn2022can}
M.~Ahn, A.~Brohan, N.~Brown, Y.~Chebotar, O.~Cortes, B.~David, C.~Finn, C.~Fu, K.~Gopalakrishnan, K.~Hausman, \emph{et~al.}, ``Do as i can, not as i say: Grounding language in robotic affordances,'' \emph{arXiv preprint arXiv:2204.01691}, 2022.

\bibitem{singh2023progprompt}
I.~Singh, V.~Blukis, A.~Mousavian, A.~Goyal, D.~Xu, J.~Tremblay, D.~Fox, J.~Thomason, and A.~Garg, ``Progprompt: Generating situated robot task plans using large language models,'' in \emph{2023 IEEE International Conference on Robotics and Automation (ICRA)}.\hskip 1em plus 0.5em minus 0.4em\relax IEEE, 2023, pp. 11\,523--11\,530.

\bibitem{huang2023grounded}
W.~Huang, F.~Xia, D.~Shah, D.~Driess, A.~Zeng, Y.~Lu, P.~Florence, I.~Mordatch, S.~Levine, K.~Hausman, \emph{et~al.}, ``Grounded decoding: Guiding text generation with grounded models for embodied agents,'' \emph{Advances in Neural Information Processing Systems}, vol.~36, pp. 59\,636--59\,661, 2023.

\bibitem{pmlr-v162-huang22a}
W.~Huang, P.~Abbeel, D.~Pathak, and I.~Mordatch, ``Language models as zero-shot planners: Extracting actionable knowledge for embodied agents,'' in \emph{Proceedings of the 39th International Conference on Machine Learning}, ser. Proceedings of Machine Learning Research, K.~Chaudhuri, S.~Jegelka, L.~Song, C.~Szepesvari, G.~Niu, and S.~Sabato, Eds., vol. 162.\hskip 1em plus 0.5em minus 0.4em\relax PMLR, 17--23 Jul 2022, pp. 9118--9147.

\bibitem{li2022pre}
S.~Li, X.~Puig, C.~Paxton, Y.~Du, C.~Wang, L.~Fan, T.~Chen, D.-A. Huang, E.~Aky{\"u}rek, A.~Anandkumar, \emph{et~al.}, ``Pre-trained language models for interactive decision-making,'' \emph{Advances in Neural Information Processing Systems}, vol.~35, pp. 31\,199--31\,212, 2022.

\bibitem{huang2022inner}
W.~Huang, F.~Xia, T.~Xiao, H.~Chan, J.~Liang, P.~Florence, A.~Zeng, J.~Tompson, I.~Mordatch, Y.~Chebotar, \emph{et~al.}, ``Inner monologue: Embodied reasoning through planning with language models,'' \emph{arXiv preprint arXiv:2207.05608}, 2022.

\bibitem{vemprala2024chatgpt}
S.~H. Vemprala, R.~Bonatti, A.~Bucker, and A.~Kapoor, ``Chatgpt for robotics: Design principles and model abilities,'' \emph{Ieee Access}, 2024.

\bibitem{song2023llm}
C.~H. Song, J.~Wu, C.~Washington, B.~M. Sadler, W.-L. Chao, and Y.~Su, ``Llm-planner: Few-shot grounded planning for embodied agents with large language models,'' in \emph{Proceedings of the IEEE/CVF international conference on computer vision}, 2023, pp. 2998--3009.

\bibitem{ajay2023compositional}
A.~Ajay, S.~Han, Y.~Du, S.~Li, A.~Gupta, T.~S. Jaakkola, J.~B. Tenenbaum, L.~P. Kaelbling, A.~Srivastava, and P.~Agrawal, ``Compositional foundation models for hierarchical planning,'' in \emph{Thirty-seventh Conference on Neural Information Processing Systems}, 2023.

\bibitem{driess2023palm}
D.~Driess, F.~Xia, M.~S. Sajjadi, C.~Lynch, A.~Chowdhery, A.~Wahid, J.~Tompson, Q.~Vuong, T.~Yu, W.~Huang, \emph{et~al.}, ``Palm-e: An embodied multimodal language model,'' 2023.

\bibitem{song2024code}
D.~Song, H.~Guo, Y.~Zhou, S.~Xing, Y.~Wang, Z.~Song, W.~Zhang, Q.~Guo, H.~Yan, X.~Qiu, \emph{et~al.}, ``Code needs comments: Enhancing code llms with comment augmentation,'' \emph{arXiv preprint arXiv:2402.13013}, 2024.

\bibitem{Ouyang_2025}
\BIBentryALTinterwordspacing
S.~Ouyang, J.~M. Zhang, M.~Harman, and M.~Wang, ``An empirical study of the non-determinism of chatgpt in code generation,'' \emph{ACM Transactions on Software Engineering and Methodology}, vol.~34, no.~2, p. 1–28, Jan. 2025. [Online]. Available: \url{http://dx.doi.org/10.1145/3697010}
\BIBentrySTDinterwordspacing

\end{thebibliography}

\end{document}